\documentclass[conference]{IEEEtran}
\IEEEoverridecommandlockouts
\usepackage{cite}
\usepackage{amsmath,amssymb,amsfonts}
\usepackage{algorithmic}
\usepackage{graphicx}
\usepackage{textcomp}
\usepackage{subfigure}
\usepackage{xcolor}
\usepackage[linesnumbered,ruled,vlined]{algorithm2e}
\def\BibTeX{{\rm B\kern-.05em{\sc i\kern-.025em b}\kern-.08em
    T\kern-.1667em\lower.7ex\hbox{E}\kern-.125emX}}
\begin{document}

\title{MRHER: Model-based Relay Hindsight Experience Replay for Sequential Object Manipulation Tasks with Sparse Rewards\\
\thanks{The research is supported by the National Science Foundation of China under Grant No.62273096, the Key Project of Dongguan Science and Technology of Social Development Program under Grant No.20211800904692 and No.20211800904492, Dongguan Sci-tech Commissoner Program under Grant No.20221800500012, Guangdong Provincial Basic and Applied Basic Research Fund Regional Joint Fundt under Grant No. 2020B1515120095, Guangdong Province Characteristic Innovation Projects of Colleges and Universities under Grant No.2022KTSCX218.}
}

\author{\IEEEauthorblockN{1\textsuperscript{st} Yuming Huang}
\IEEEauthorblockA{\textit{School of Computer Science and Technology} \\
\textit{Dongguan University of Technology}\\
Dongguan, China \\
hym\_troy@qq.com}
\\
\IEEEauthorblockN{3\textsuperscript{rd} Ziming Xu}
\IEEEauthorblockA{\textit{International School of Microelectronics}\\
\textit{Dongguan University of Technology}\\
Dongguan, China \\
505642573@qq.com}
\and
\IEEEauthorblockN{2\textsuperscript{nd}  Bin Ren*}
\IEEEauthorblockA{\textit{International School of Microelectronics}\\
\textit{Dongguan University of Technology}\\
Dongguan, China \\
renbin@dgut.edu.cn}
\\
\IEEEauthorblockN{4\textsuperscript{th} Lianghong Wu}
\IEEEauthorblockA{\textit{International School of Microelectronics}\\
\textit{Dongguan University of Technology}\\
Dongguan, China \\
wulianghong2022@email.szu.edu.cn}
}

\maketitle

\begin{abstract}
Sparse rewards pose a significant challenge to achieving high sample efficiency in goal-conditioned reinforcement learning (RL). Specifically, in sequential manipulation tasks, the agent receives failure rewards until it successfully completes the entire manipulation task, which leads to low sample efficiency.
To tackle this issue and improve sample efficiency, we propose a novel model-based RL framework called Model-based Relay Hindsight Experience Replay (MRHER). MRHER breaks down a continuous task into subtasks with increasing complexity and utilizes the previous subtask to guide the learning of the subsequent one.
Instead of using Hindsight Experience Replay (HER) in every subtask, we design a new robust model-based relabeling method called Foresight relabeling (FR). FR predicts the future trajectory of the hindsight state and relabels the expected goal as a goal achieved on the virtual future trajectory. By incorporating FR, MRHER effectively captures more information from historical experiences, leading to improved sample efficiency, particularly in object-manipulation environments.
Experimental results demonstrate that MRHER exhibits state-of-the-art sample efficiency in benchmark tasks, outperforming RHER by 13.79\% and 14.29\% in the FetchPush-v1 environment and FetchPickandPlace-v1 environment, respectively.
\end{abstract}


\section{Introduction}
\label{sec:intro}

In reinforcement learning (RL), the training of agents always requires a well-designed shaped reward to optimize the policy. Particularly in sequential manipulation tasks, designing a well-worked shaped reward requires professional knowledge and cannot be easily transferred to another task. So, sparse reward is introduced into the training of agents \cite{durugkar_adversarial_2021,bing_solving_2023,chen_imitation_2023,luo_relay_2022}, which indicates whether the agents complete the task or not by using a binary reward signal. Our paper focuses on how to improve the sample efficiency in sequential manipulation tasks with sparse rewards.

In sparse-rewards tasks, agents are difficult to distinguish the effectiveness of the current actions because the negative rewards are given to the actions before the expected goals are achieved. However, agents can effectively optimize the policy and learn well only when a sufficient number of samples with non-negative rewards are collected.
\begin{figure}[]
\centering
\includegraphics[width=1\linewidth]{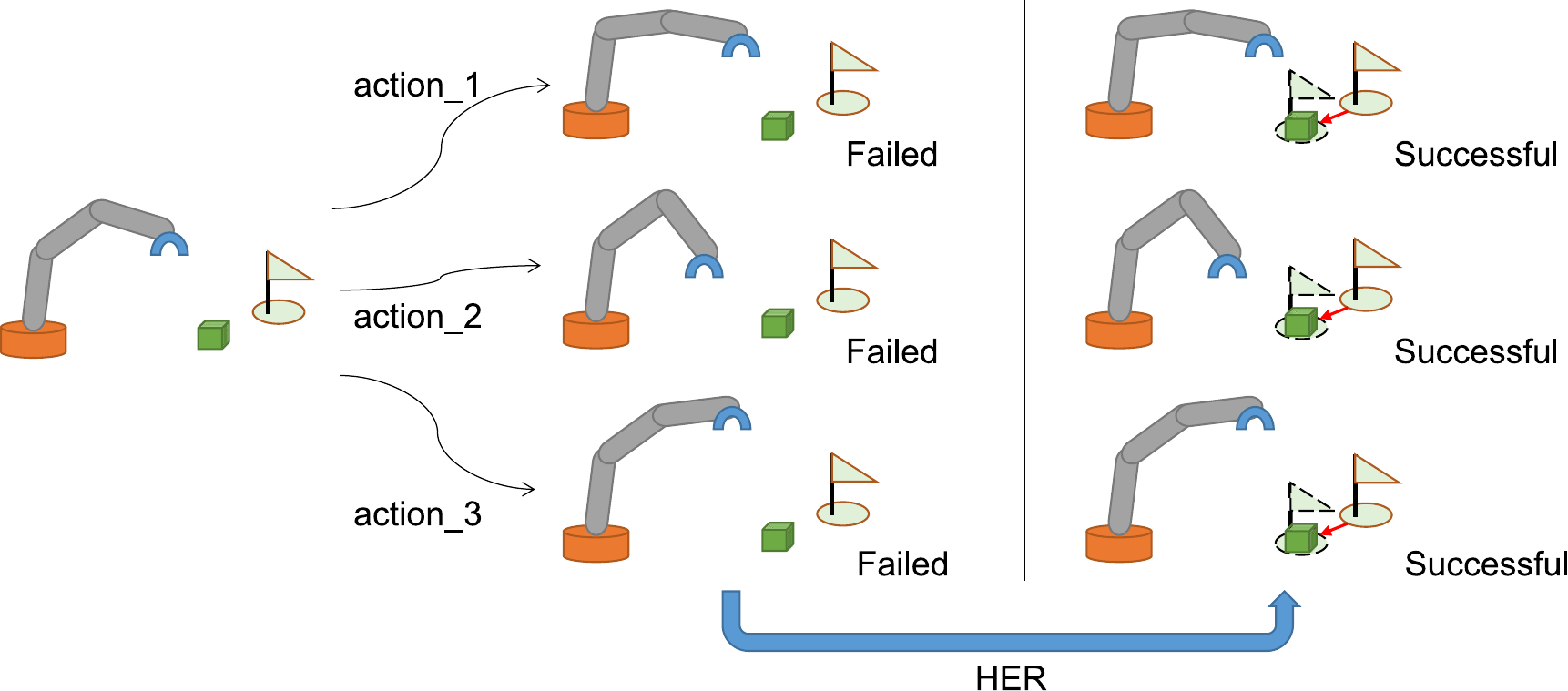}
\caption{The Identical Non-Negative Reward (INNR) problem of HER.}
\label{fig:1.0}
\end{figure}

Hindsight Experience Replay (HER) \cite{andrychowicz_hindsight_2017} is one of the effective goal relabeling methods to produce more samples with non-negative rewards. Inspired by the hindsight ability of humans, HER replaces expected goals by achieved goals in failure trajectories and recalculates the rewards into non-negative rewards. However in sequential manipulation tasks, the agents suffer from the Identical Non-Negative Reward (INNR) problem of HER during the exploration process \cite{Manela_Biess_2021, Manela_Biess_2022, luo_relay_2022}. When the actions of agents do not affect achieved goals, if HER relabels the goals and give the non-negative rewards to all actions, the INNR problem occurs and disables agents to differentiate the value of different actions and to effectively optimal the policy, as shown in Fig.\ref{fig:1.0}. Based on HER, Model-based relabeling (MBR) \cite{zhu_mapgo_2021,yang_mher_2021}  introduces dynamics models to predict the future trajectories of the current transitions for goal relabeling, the INNR problem arises due to the high requirement for accurate prediction of the dynamics models, limiting the widespread use of MBR in sequential manipulation tasks. More details are discussed in Section \ref{4.1}.


To alleviate the INNR problem and improve the sample efficiency, we propose a model-based framework, named Model-based Relay Hindsight Experience Replay (MRHER). MRHER breaks down a continuous task into incremental subtasks and utilizes the previous subtask to guide the subsequent one. Instead of using HER in subtasks, we design a new model-based relabeling method called Foresight relabeling (FR), which uses the newest policy interacting with trained dynamics models to obtain future trajectories for relabeling goals, as shown in Fig.\ref{fig:1.1}. Modifying the expected goal to a pseudo goal can be assumed that the agent uses the newest policy and a different goal to reinterpret its actions, resulting in policy-guided and goal-relabeling implicit courses \cite{yang_mher_2021}. 

The primary contributions of our research are outlined in the following:
\begin{itemize}
\item To improve the sample efficiency in sparse-rewards sequential manipulation tasks, this paper develops a model-based and sample-efficient RL framework called MRHER.
\item To alleviate the INNR problem caused by the errors of dynamics models and accelerate the exploration, we design a model-based relabeling method whose dynamics models are designed to act on hindsight $future$ states with the newest subsequent policy.
\item To evaluate the performance of MRHER, we conducted experiments in four robot sequential manipulation environments. Experimental results show that MRHER achieves higher sample efficiency compared to HER and other state-of-the-art baselines.
\end{itemize}

\section{Related Work}
\label{sec:formatting}
The main focus of our work is to enhance the sample efficiency of sparse-reward goal-conditioned tasks, which is related to HER-based methods and Model-based RL.

\paragraph{HER-Based Methods}
Inspired by the ability of humans to learn from failure experiences, Hindsight Experience Replay (HER) \cite{andrychowicz_hindsight_2017} was proposed in sparse-reward Goal-Conditioned RL (GCRL), which relabels the expected goals and enables the agent to learn from failure trajectories. HER is effective in long-horizon reaching tasks \cite{andrychowicz_hindsight_2017}, but not in sequential manipulation tasks \cite{Manela_Biess_2022}. 

To improve the sample efficiency of HER, a number of HER-based methods have been proposed. Curriculum-guided HER (CHER) \cite{fang_curriculum-guided_2019} selects appropriate goals for relabeling by dynamically balancing goal proximity and diversity. To address the long horizon problems, Maximum Entropy Goal Achievement (MEGA) \cite{kuang_goal_2020} optimizes the expected goals to encourage agents to explore sparse areas of the goal space, particularly the edges of the space. Filtered-HER \cite{Manela_Biess_2021} was the first to identify the issue of INNR and utilizes a filter to remove misleading samples. Different from Filtered-HER, Sequential-HER (SHER) \cite{Manela_Biess_2022} breaks down a continuous task into a set of sequential subtasks and transfers the previous policy to the next one to solve the INNR problem. To achieve self-guided exploration, Relay Hindsight Experience Replay (RHER) \cite{luo_relay_2022} designs a multi-goal \& multi-subtasks network to learn multiple subtasks simultaneously and utilize the previous policy to guide the next one. However RHER learns without dynamics models, so it cannot improve its sample efficiency by predict future trajectories. 

\paragraph{Model-Based RL}
Compared to model-free RL algorithms, model-based RL algorithms always exhibit higher sample efficiency \cite{atkeson_comparison_1997}. Inspired by humans' ability to predict future trajectories, Dyna \cite{sutton_dyna_1991} was the first model-based RL algorithm to increase training data by generating virtual samples with a learned dynamics model. 
Trained dynamics models can be used to predict into the future to improve the policy learning  and value estimation of a model-free algorithm \cite{nagabandi_neural_2018}. Model-based value expansion (MVE) \cite{Feinberg_2018} method introduces the predictive transitions to improve the value function estimation. To mitigate the impact of model bias in learned models, STEVE \cite{buckman2018sample} uses model ensembles for a robust prediction. Model-Based Policy Optimization (MBPO) proves shorter predictive rollout can reduce the effects of model inaccuracies. However, most model-based RL algorithms are designed for dense-rewards tasks and are not suitable for sparse-rewards tasks. G-HER \cite{bai_guided_2019} firstly uses a conditional generative recurrent neural network (RNN) to explicitly model the relationship between policy level and goals, generating a goal with a high average return value under the current policy for relabeling. PlanGAN \cite{charlesworth_plangan_2020} combines GANs and models for planning. But, one drawback of PlanGAN is that the process of simulating trajectories for action selection requires a significant amount of computation, and the algorithm may accumulate model errors over long planning horizons. Model-Assisted Policy Optimization for Goal-Oriented Tasks (MapGo) \cite{zhu_mapgo_2021} and Model-based Hindsight Experience Replay (MHER) \cite{yang_mher_2021} firstly use dynamics models to predict future trajectories for model-based goal relabeling. MapGo, using visual actions and states to train the policy, results in a substantial amount of models' error. MHER avoids virtual actions or states by combining policy gradient and Goal-Conditioned Supervised Learning (GCSL) \cite{ghosh_learning_2021} to update the policy. However, both methods cause the INNR problem because of the dynamics models' errors in sequential manipulation environments. To alleviate the INNR problem and effectively utilize dynamic models in such environments, MRHER owns a new model-based relabeling method to improve sample efficiency.

\section{Preliminaries}
\subsection{Goal-Conditioned Reinforcement Learning}
Goal-Conditioned reinforcement learning (GCRL) can be mathematically represented by a tuple \textless $S, A, G, r, \mathcal{T}, \gamma$ \textgreater
, where $S, A, G, \mathcal{T}, \gamma$ denote the state space, action space, goal space, transition function, and discount factor, respectively. The reward function $r: S \times A \times G \to \mathbb{R}$ is goal-conditioned in sparse-rewards environments \cite{schaul_universal_2015}, and is defined as: 
\begin{equation}
\label{eq1}
r\left(s_{t}, a_{t}, g\right)=\left\{
\begin{array}{ll}
0, & \left\|\phi\left(s_{t}\right)-g\right\|_{2}^{2}<\mathrm{threshold} \\
-1, & \mathrm { otherwise }
\end{array}\right.
\end{equation}

where, $\phi: S \to G$ is a tractable function that maps states to achieved goals \cite{andrychowicz_hindsight_2017}. 
The objective of GCRL is to optimize a policy that maximizes the cumulative reward of the goal distribution, which can be expressed as:
\begin{equation}
\label{eq2}
J(\pi)=E_{g \sim p(g), a_{t} \sim \pi, s_{t+1} \sim \mathcal{T}\left(\cdot \mid s_{t}, a_{t}\right)}\left[\sum_{t=0}^{\infty} \gamma^{t} r\left(s_{t}, a_{t}, g\right)\right]
\end{equation}

\subsection{Hindsight Experience Replay}
One of the core methods for addressing the problem of sparse rewards in GCRL is Hindsight Experience Replay (HER), which can be integrated into any off-policy RL algorithm \cite{mnih_human-level_2015,lillicrap_continuous_2015,haarnoja_soft_2018}. The idea of HER is to modify the expected goals of failed experiences to achieved goals, which allows for reevaluating the value of current actions from different goals. At the start of each trajectory, the agent is given a goal $g$ and generates a trajectory $\tau = {(s_t, a_t, s_{t+1}, g, r_t)}_{t=1}^T $ of length $T$ with $\pi$. Then the trajectory is saved in the replay buffer. For the t-th transition tuple $(s_t, a_t, s_{t+1}, g, r_t)$, HER relabels the expected goal $g$ with a achieved goal $g' = \phi(s)$ mapped by the state.
HER employs three commonly used strategies: $future$, $episode$, and $final$. The $future$ strategy is the most frequently used strategy, which relabels the expected goal $g$ as a state after $s_t$ in the same trajectory; the $episode$ strategy selects an arbitrary state in the same trajectory to relabel the expected goal $g$; the $final$ strategy relabels $g$ with the last state $s_T$ in the same trajectory. Finally, the reward is recalculated using \eqref{eq1} as $r_t' = (s_t, a_t, g')$.

\section{Methodology}
\label{4}
In this section, we present MRHER, a robust framework that combines dynamics models with Hindsight Experience Replay (HER) \cite{andrychowicz_hindsight_2017}. We first discuss the motivation behind our approach, followed by explanations of decomposition, recombination and self-guided exploration. Then, we describe how to utilize historical experience to train dynamics models capable of predicting future trajectories. Next, we propose a new relabeling method called Foresight Relabeling based on trained models,. Finally, we provide a comprehensive overview of the overall framework of MRHER.

\begin{figure}[!t]
\centering
\includegraphics[width=1\linewidth]{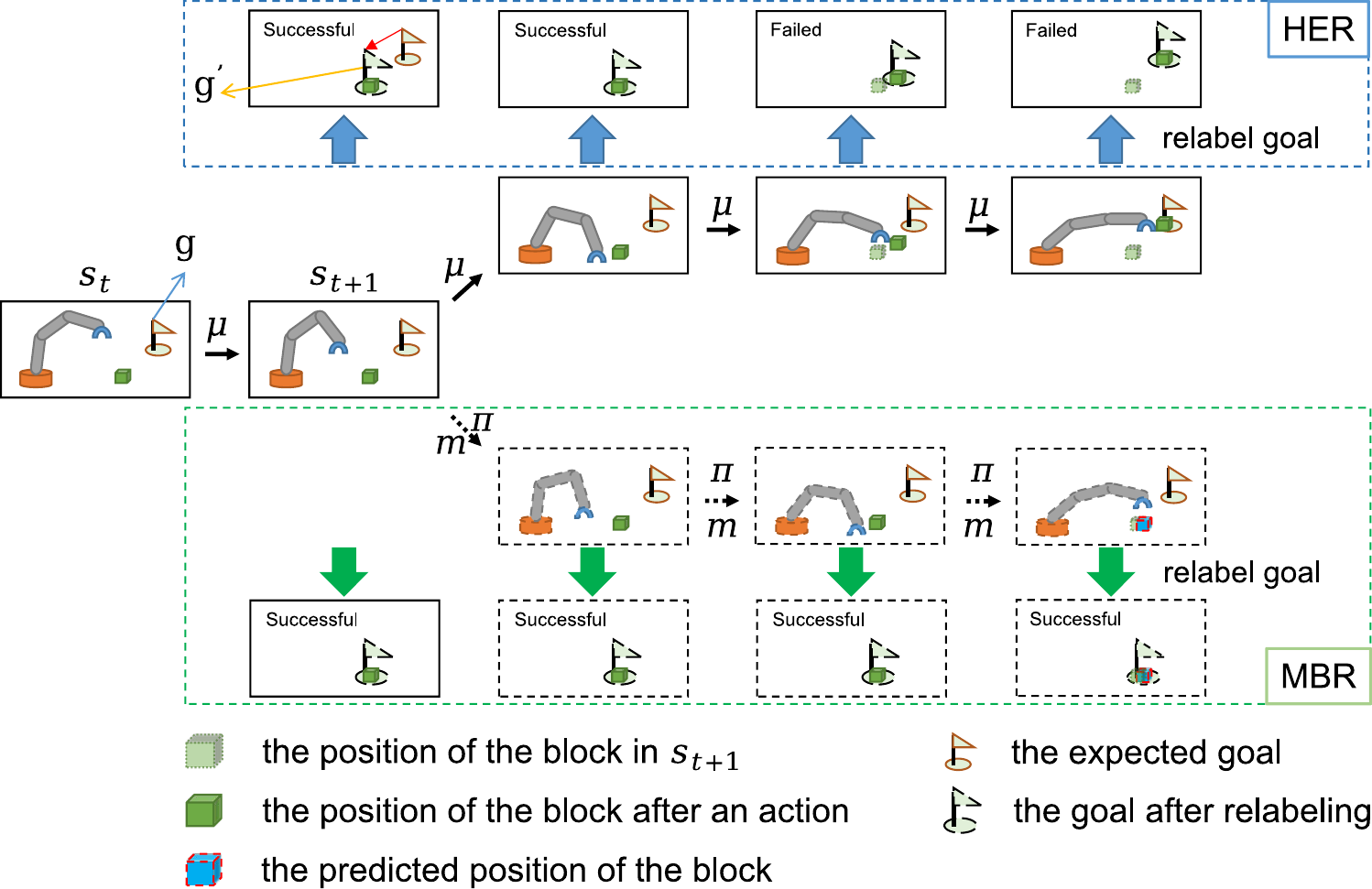}
\caption{The INNR problem of Model-based Relabeling. Real trajectories are collected by the agent based on the past policy $\mu$, while virtual trajectories are collected based on the newest policy $\pi$ and dynamical models. HER uses the achieved goals in the historical trajectories to relabel the expected goals. MBR predict the future trajectories of the current transition for goal relabeling.}
\label{fig:4.1}
\end{figure}
\subsection{Motivation: The INNR Problem of Model-based Relabeling}
\label{4.1}
In sequential manipulation tasks, agents need to accomplish a series of manipulation to achieve the expected goal. When HER are applied in these tasks, the INNR problem occurs \cite{Manela_Biess_2021, Manela_Biess_2022, luo_relay_2022}. For a block-pushing task, the agent should reach the position of the block and push it to an expected position. As shown in Fig.\ref{fig:1.0}, HER considers the block's position as the achieved goal $g'$, relabels the expected goal $g$ as $g'$ and recalculate a non-negative reward $r'$. So, although the actions of an agent do not move the block, this kind of non-negative rewards create an illusion that all actions are good and disables the agent to differentiate the value of different actions. This phenomenon is called the INNR problem. 

Based on HER, model-based relabeling (MBR) \cite{zhu_mapgo_2021,yang_mher_2021} methods are proposed. MBR trains the dynamics models $m$ with historical trajectories and utilizes $m$ and the newest policy $\pi$ to predict the $n$-step future trajectories of the current transition for goal relabeling. However, MBR only choose the state $s_{t+1}$ of the current transition as the beginning of virtual future trajectories, that leads to an even more serious INNR problem. In the block-pushing task, most of the agent's trajectories in early exploration do not affect the position of the block and are collected to train the dynamics models $m$, resulting in a tendency for $m$ to ignore movements of the block in predicted future trajectories. This means that MBR relabels the expected goal $g$ as a achieved goal $g'$ on the virtual trajectory, which is equivalent to relabeling $g$ as the achieved goal $g'$ mapped by the state $s_{t+1}$ and recomputing the non-negative reward $r_t' = (s_t, a_t, g'=\phi(s_{t+1}))$ to all actions. 

In contrast, MRHER effectively alleviates the INNR problem caused by the inaccurate models and improve the sample efficiency in sequential manipulation tasks.

\subsection{Decomposition, Recombination and Self-Guided Exploration}
\label{4.2}
To provide a foundation for the development of MRHER, we first introduce how to decompose and recombine a sequential task and to learn by self-guided exploration.

\begin{figure}[!t]
\centering
\includegraphics[width=1\linewidth]{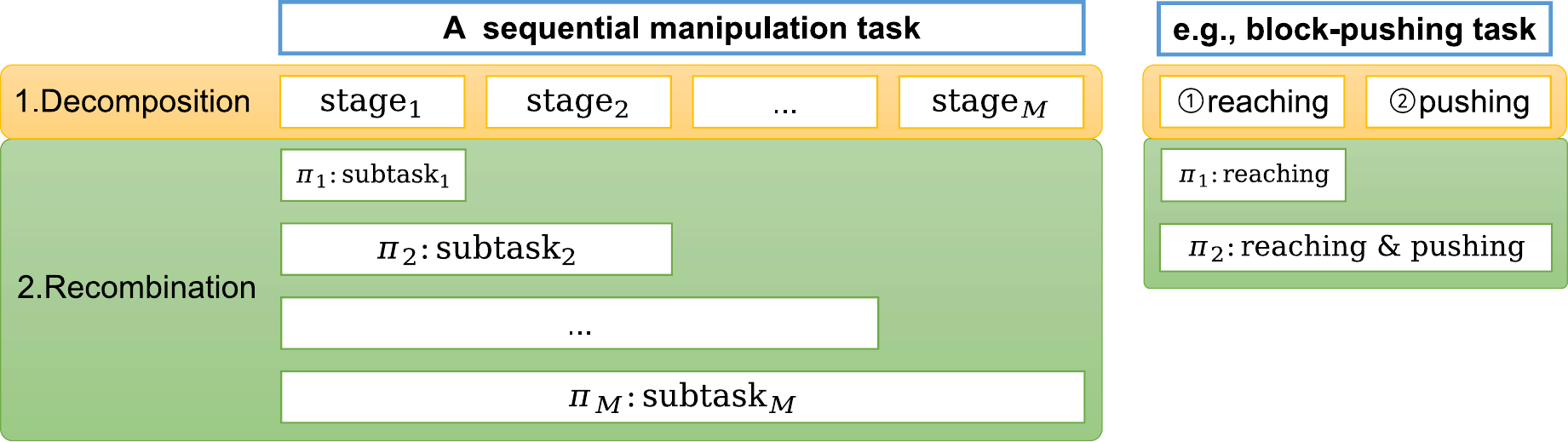}
\caption{The decomposition and recombination of a sequential task.}
\label{fig:4.2}
\end{figure}
\textbf{Decomposition and Recombination:} Similar to the ability of humans to divide a sequential manipulation task into several ordered stages, MRHER decomposes a task into multiple stages and recombines them into subtasks of increasing complexity, as shown in Fig.\ref{fig:4.2}. Given that $N$ objects need to be manipulated in a task, the task can be decomposed into $M=2N$ stages because the agent must first reach an object $i$ and then manipulate it. If the goal of a subtask has been achieved, the agent progresses to the next subtask and the goal changes at the same time. So, the index $j$ of the current stage is determine by (\ref{eq4.2}):
\begin{equation} 
\label{eq4.2}
j = \begin{cases} 2 \times i-1 & \text{if } dist_{i} < d \\ 2 \times i & \text{if } dist_{i} \geq d \end{cases} 
\end{equation}
where $d$ is a defined distance threshold and $dist_i$ represents the distance between the agent and the $i$-th object. 

After decomposing and recombining the task, the agent starts with the simple subtasks and progressively guides the exploration of more complex subtasks.

\textbf{Self-Guided Exploration Strategy:} To efficiently use a simple subtask to guide the next complex subtask, MRHER mixes the policy of complex subtask with the policy of simple subtask, which is called Self-Guided Exploration Strategy (SGES) \cite{luo_relay_2022}. The policy of simple subtask that has been trained well to finish the simple subtask is treated as the already-learned policy. Meanwhile, the policy of the complex subtask, that needs to achieve the goal of simple subtask first and then explore next stage, is considered as the learning policy. When the learning policy tries to finish the simple subtask, the already-learned policy is activated with a certain probability to take actions. If the learning policy takes incorrect or suboptimal actions in the simple stage, the already-learned policy can help correct these actions in time. With the help of the already-learned policy, the learning policy could efficiently finish the simple subtask, transition to the new goal and collect valuable samples of the next stage.
\begin{figure}[!t]
\centering
\includegraphics[width=1\linewidth]{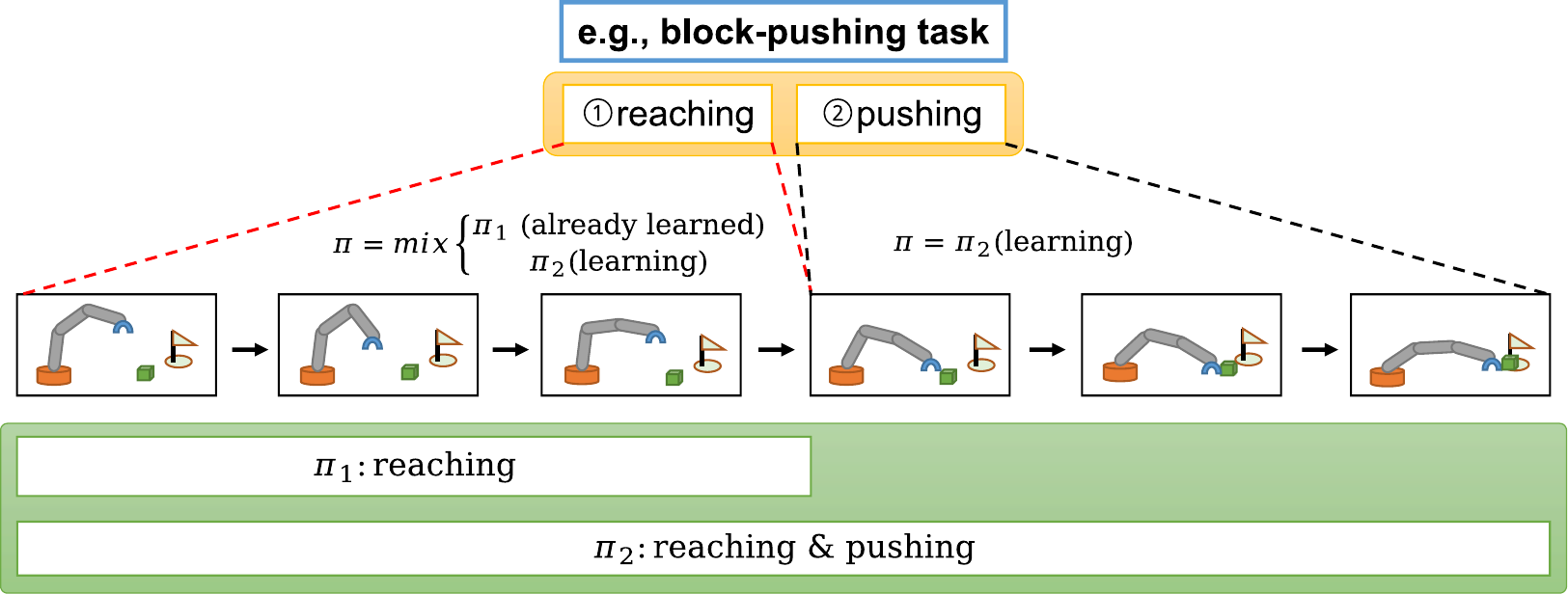}
\caption{The SGES in a block-pushing task.}
\label{fig:4.3}
\end{figure}

For a block-pushing task, MRHER decomposes the whole task into two stages: a reaching stage and a pushing stage, as shown in Fig.\ref{fig:4.3}. These stages are then recombined into two subtasks: a reaching subtask and a reaching \& pushing subtask.
In the reaching subtask, the agent's goal is to reach the target block without considering the pushing action. To achieve this goal efficiently, the agent learns a reaching policy, denoted as $\pi_1$. Once the reaching subtask is completed, the agent progresses to the reaching \& pushing subtask. 
In the reaching \& pushing subtask, the goal is to reach the target block and push it to the expected location. The agent learns a policy $\pi_2$ of the subtask (including reaching and pushing stages) to accomplish the goal. When $\pi_2$ is learning the reaching skill, MRHER mixes the already-learned reaching policy $\pi_1$ and the learning policy $\pi_2$ to expedite the learning process of the reaching skill. With the help of $\pi_1$, $\pi_2$ can quickly learn the reaching skill well and explore the pushing stage.

To alleviate the INNR problem of MBR, MRHER decomposes a sequential task, recombines it into subtasks and utilizes SGES to finish the simple subtasks quickly and to collect valuable samples of the complex subtasks.

\subsection{Dynamics Models}
In MRHER, the integration of dynamics models plays a crucial role in improving the agent's understanding of environment dynamics and enhancing sample efficiency. In sequential manipulation tasks, directly using a model $m(s_t, a_t)$ to map $s_t$ and $a_t$ to $s_{t+1}$ will lead to the action $a_t$ being ignored due to the similarity between adjacent states $s_t$ and $s_{t+1}$ \cite{nagabandi_neural_2018}. Compared to directly predicting the next state, we follow \cite{nagabandi_neural_2018, yang_mher_2021} and learn to predict the differences between states using models and compute the next state as $s_{t+1} = s_t + m(s_t, a_t)$. To train the dynamics models, we minimize the following error formula:
\begin{equation}
\mathcal{L}_{\text {model}}=E_{\left(s_{t}, a_{t}, s_{t+1}\right) \sim B}\left[\left\|\left(s_{t+1}-s_{t}\right)-m\left(s_{t}, a_{t}\right)\right\|_{2}^{2}\right]
\label{eq3}
\end{equation}
where, relabeled data can be directly used to train the models since the data $(s_t, a_t, s_{t+1})$ does not involve the target and reward. 
After training dynamics models, the agent gains the capability to predict states through interaction with the models.

\begin{figure}[t]
\centering
\includegraphics[width=1\linewidth]{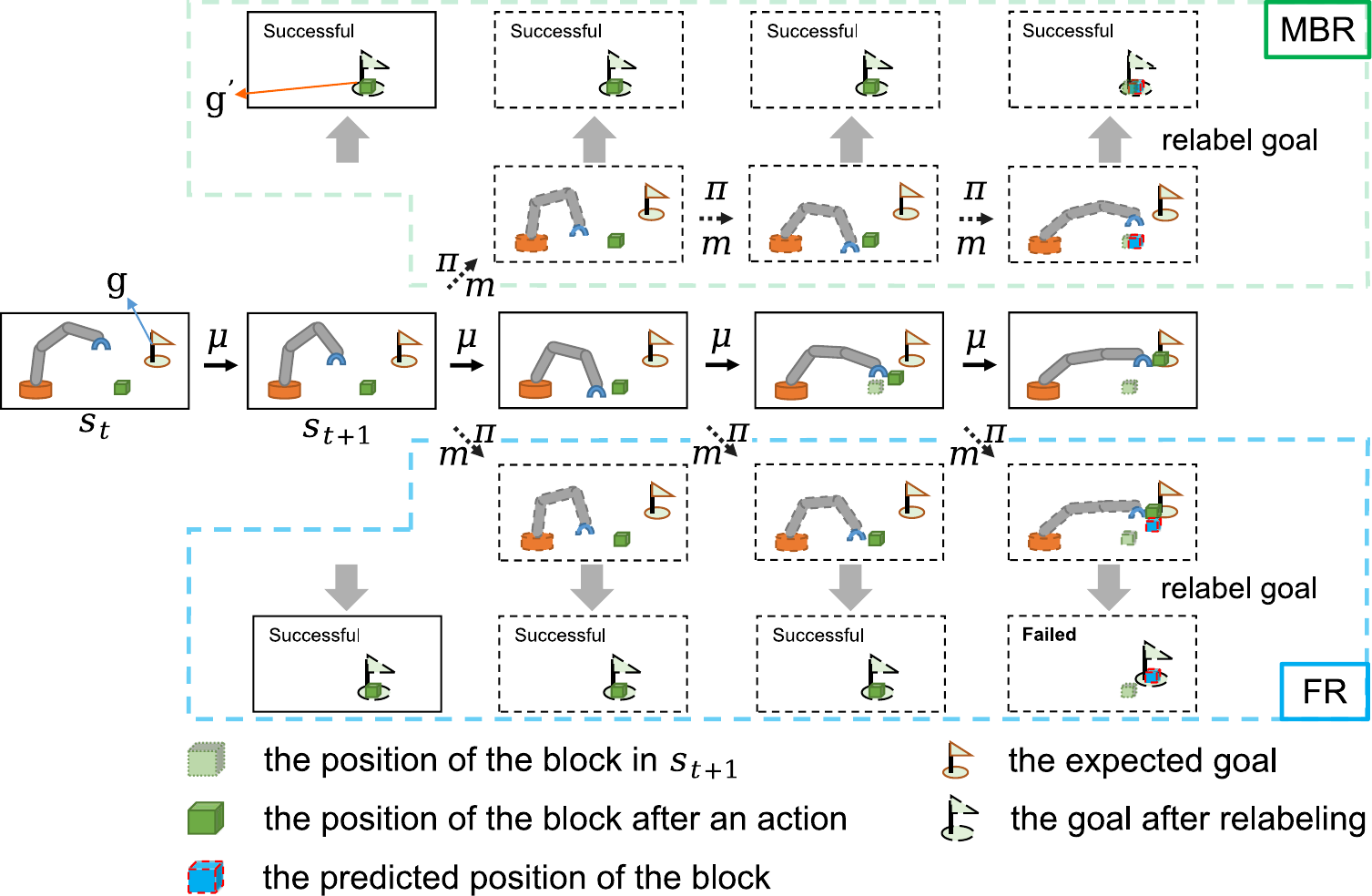}
\caption{Diagram of Foresight relabeling. Foresight relabeling selects a beginning state, generates a virtual trajectories, and then uses the achieved goals on the virtual trajectories to relabel the expected goal.}
\label{fig:1.1}
\end{figure}

\begin{algorithm}[t]
    \caption{Foresight relabeling}
    \label{alg4.4}
    $\mathbf{Iuput:}$ minibatch $b$, step $n$, dynamics models $m$\;
   
    
    \For{$\left(s_{t}, a_{t}, s_{t+1}, g, r_{t}\right)$ in $b$ }
    {        
        Initialize a virtual trajectory $\tau^{\prime} \gets \emptyset$ \;
        Sample a state $s_{k}$ on the same trajectory, $t+1\leq k \leq T$\;
        $s_{k}^{\prime}=s_{k},\tau^{\prime}=\{s_{k}^{\prime}\}$\;
         \For{$l$ = 0, 1, . . . , n-1}
        {
            $a_{k+l}^{\prime}=\pi\left(s_{k+l}^{\prime}, g\right)$ \\            
            $s_{k+l+1}^{\prime}=s_{k+l}^{\prime}+m\left(s_{k+l}^{\prime}, a_{k+l}^{\prime}\right)$ \\ 
            Append $s_{k+l+1}^{\prime}$ to $\tau^{\prime}$ \\
        }
        Update $\tau^{\prime}$ in (\ref{eq4.4})\;
        Choose a state $s^{\prime} \sim \tau^{\prime}$\;
        Get the virtual achieved goal $g^{\prime}=\phi(s^{\prime})$\;
        Calculate the new reward $r_{t}^{\prime} = r(s_{t}, a_{t}, g^{\prime})$ in \eqref{eq1}\;
        Store the transition $\left(s_{t}, a_{t}, s_{t+1}, g^{\prime}, r_{t}^{\prime}\right)$ in $b$\;
    }
    return $b$;
\end{algorithm}

\subsection{Foresight Relabeling}
With dynamics models, we propose a new relabeling method called Foresight relabeling (FR) to address the INNR problem of MBR and further improve the performance of MRHER.

Similar to MBR, FR also predicts future trajectories for goal relabeling.
However, instead of selecting $s_{t+1}$ of the current transition as the beginning of future trajectories, utilizing different historical states as the beginning can avoid relabeling all expected goals as the achieved goal of $s_{t+1}$ and alleviate the INNR problem. As the Fig.\ref{fig:1.1} shows, given a transition $(s_t, a_t, s_{t+1}, g, r_t)$, a beginning state $s_k$ after the state $s_t$ on the same trajectory is randomly selected. Then the agent interacts with the dynamical models $m$ using the newest policy $\pi$ to predict a virtual $n$-steps trajectory $\tau'$ starting from $s_k$. Finally, FR chooses a state $s^{\prime}$ from $\tau'$, relabels the expected goal $g$ to a new goal $g' = \phi(s^{\prime})$ and recomputes the reward $r_t'$ according to \eqref{eq1}. The transition is modified to $(s_t, a_t, s_{t+1}, g', r_t')$. In the transitions relabeled by FR, only the goal is virtual, which avoids the use of virtual states and actions to train the value function. The algorithm of FR is shown in Algo. \ref{alg4.4}.

Although different beginning states can alleviate the INNR problem of MBR, exploring a large number of virtual goals would result in lower sample efficiency due to the overlapping exploration of some highly similar goals. Additionally, as the number of predicted steps increases, the negative reward signals gradually increase while the proportion of positive reward signals decreases, which hinders the agent's ability to assess action values accurately.
To balance the goal diversity, sample efficiency, and different reward signals, the predicted goal $\phi\left(s_{k+l+1}^{\prime}\right)$ ($0 \leq l \leq n-1$) in $\tau'$ should follow the \eqref{eq4.4}, which determines the similarity of the goals by the threshold value of the reward function in \eqref{eq1}, and selects to keep or reset as $\phi\left(s_k^{\prime}\right)$ for increasing the proportion of positive reward signals.
\begin{equation} 
\label{eq4.3}
dist_{s_{k+l+1}^{\prime}, s_{k+l}^{\prime}} = \left\|\phi\left(s_{k+l+1}^{\prime}\right)-\phi\left(s_{k+l}^{\prime}\right)\right\|_{2}^{2}
\end{equation}
\begin{equation} 
\label{eq4.4}
\phi\left(s_{k+l+1}^{\prime}\right) = 
\begin{cases} \phi\left(s_{k+l+1}^{\prime}\right), & dist_{s_{k+l+1}^{\prime}, s_{k+l}^{\prime}} \geq \mathrm{threshold} \\
\phi\left(s_k^{\prime}\right), & dist_{s_{k+l+1}^{\prime}, s_{k+l}^{\prime}} < \mathrm{threshold} \end{cases} 
\end{equation}
where, $dist_{s_{k+l+1}^{\prime}, s_{k+l}^{\prime}}$ is the distance between the goals $\phi\left(s_{k+l+1}^{\prime}\right)$ and $\phi\left(s_{k+l}^{\prime}\right)$.


\begin{figure}[t]
\centering
\includegraphics[width=0.9\linewidth]{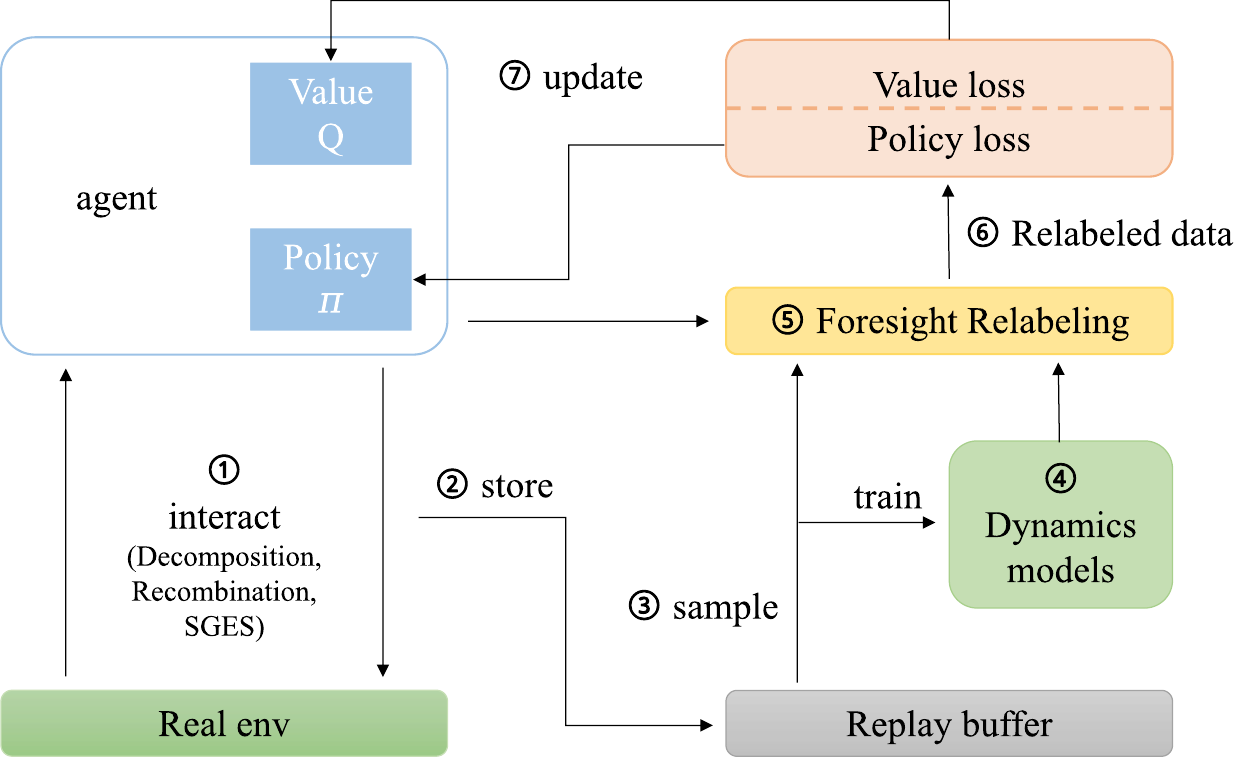}
\caption{The framework of MRHER.}
\label{fig:1}
\end{figure}

\subsection{Overall Framework of MRHER}

The MRHER framework, as illustrated in Algo. \ref{alg1} and Fig.\ref{fig:1}, follows a multi-step process. Initially, the agent interacts with the environment, starting from an initial state $s_1$, and proceeds by checking the current stage of subtasks, sampling actions using the SGES, and attempting to achieve the initial goal $g$. Subsequently, the resulting trajectory is stored in the replay buffer $B$.
Next, in each subtask, a minibatch $b_i$ of transitions is sampled from $B$ to train dynamics models $m$. For each transition in $b_i$, the Foresight Relabeling (FR) method is used. FR selects a beginning state and predicts an n-step trajectory using the models $m$ and the newest policy $\pi$ of subsequent subtasks. The achieved goal in the trajectory is then used to relabel the expected goal, and the reward is recalculated accordingly. 
Finally, the policy $\pi$ and value $Q$ are trained by the minibatch $b_i$ of relabeled transitions. In this work, the Deep Deterministic Policy Gradient (DDPG) algorithm \cite{lillicrap_continuous_2015} is adopted as the benchmark off-policy RL algorithm. 

\begin{algorithm}[t]

    \caption{MRHER Framework}
    \label{alg1}
    $\mathbf{Require:}$ off-policy RL algorithm $\mathbb{A}$, replay buffer $B$, step $n$, dynamics models $m$;
    
    Initialize $\mathbb{A}$, $B \gets \emptyset $
    \For{$episode$ = 1, 2, ..., K}
    {
         Sample an initial state $s_{0}$ and a goal $g$\;
         \For{time step $t$ = 1, 2, ..., T}{
         Check the stage via \eqref{eq4.2}\;
         Sample an action $a_{t}$ with the SGES\;
         Execute the action $a_{t}$ and obtain a new observation $s_{t+1}$\;
         }
         Store the trajectory $\left\{\left(s_{t}, a_{t}, s_{t+1}, g, r_{t}\right)\right\}_{t=1}^{T}$ in the replay buffer $B$\;
         \For{Batch = 1, 2, ..., $N_{batch}$}
         {
            \For{$subtask$ = 1, 2, ..., M}
            {
            Sample a minibatch $b$ from $B$\;
            Update dynamics models $m$ with $b$\ in \eqref{eq3}\;
            Use FR via Algo.\(\ref{alg4.4}\) to relabel $b$\;
            Optimize the policy $\pi$ and value $Q$ of $\mathbb{A}$ with $b$\;
            }
         }
    }
\end{algorithm}

\section{Experiments}

In this section, we conduct experiments on four sequential manipulation environments. Four sequential manipulation environments, baseline algorithms, and training parameter settings are introduced first. Next, the sample efficiency of MRHER and baseline algorithms is compared in the environments. Finally, the effectiveness of components in the MRHER framework are analyzed by the ablation experiments.

\subsection{Environments}
\begin{figure}[t]
  \centering
  
  \subfigure[FetchPush-v1]{
        \includegraphics[width=.40\linewidth]{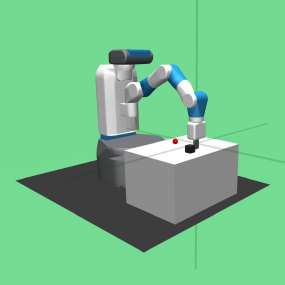}
        
        \label{fig:short-a}
    }
    \subfigure[FetchPickandPlace-v1]{
        \includegraphics[width=0.40\linewidth]{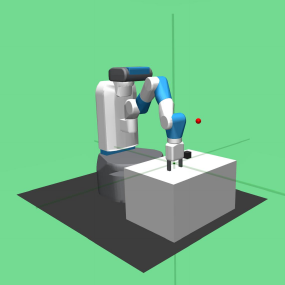}
        \label{fig:short-b}
    }
\subfigure[Drawer-v1]{
        \includegraphics[width=0.40\linewidth]{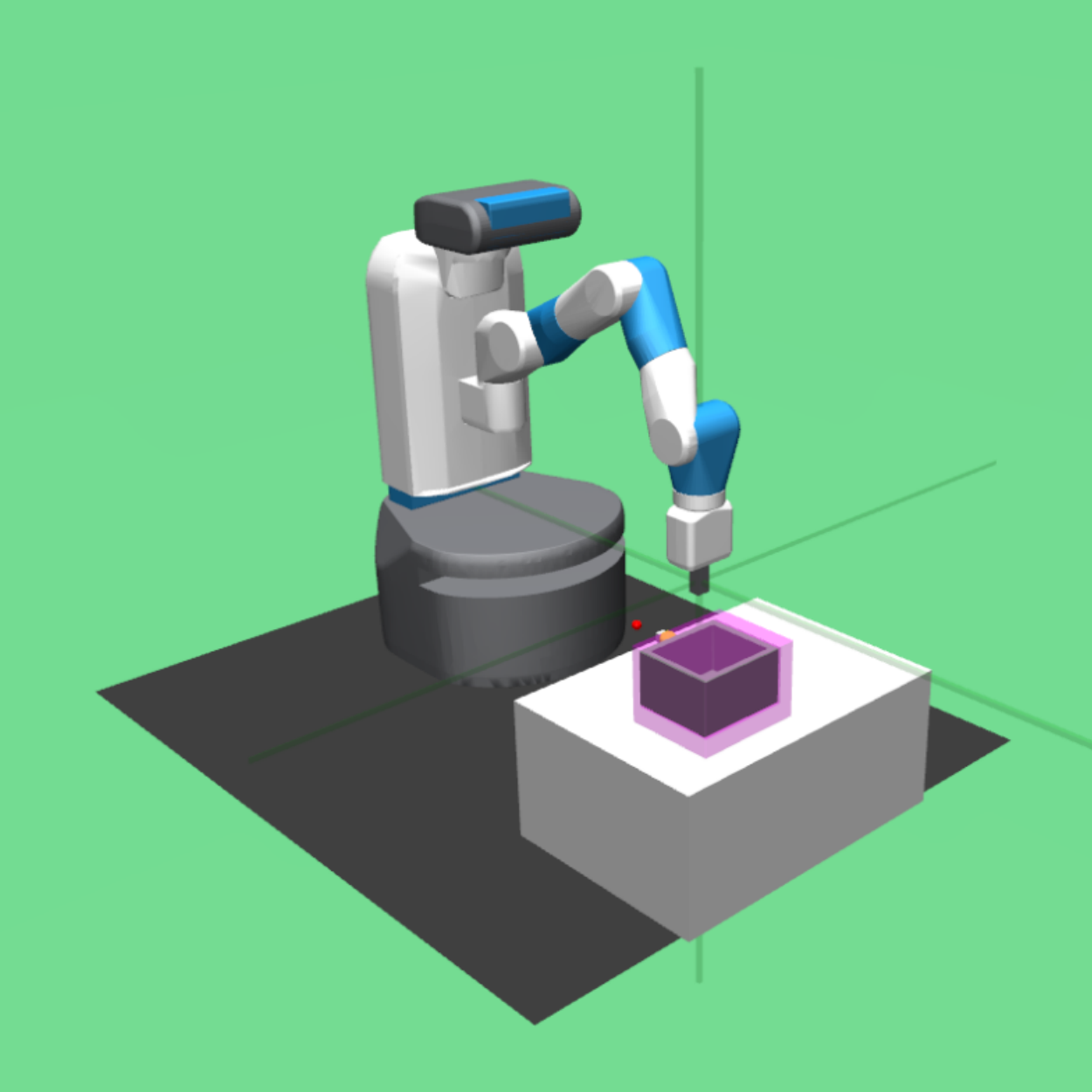}
        \label{fig:short-c}
    }
\subfigure[Insert-v1]{
        \includegraphics[width=0.40\linewidth]{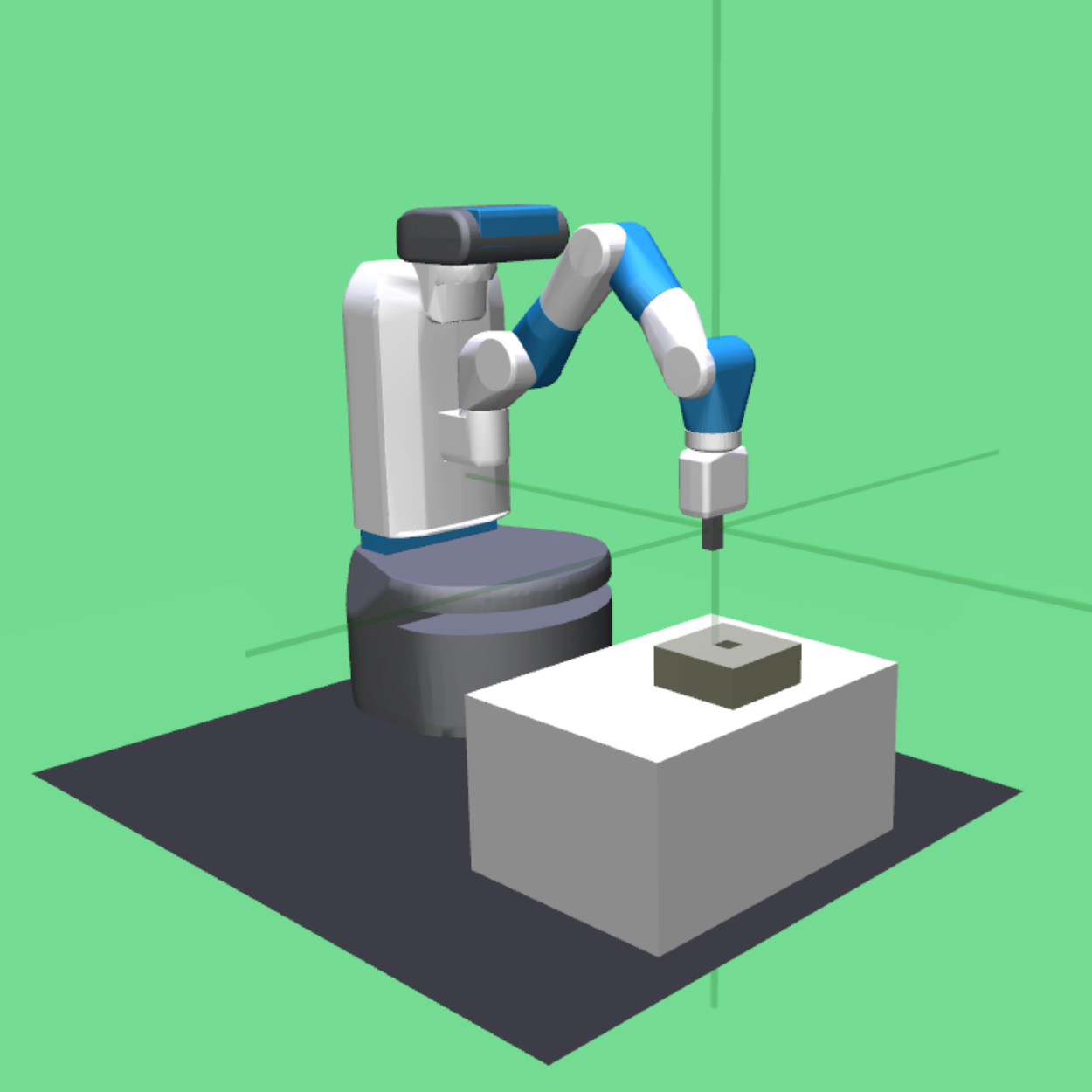}
        \label{fig:short-d}
    }

  \caption{Different sequential object manipulation tasks.}
  \label{fig:2}
\end{figure}

Fig.\ref{fig:2} shows four sequential manipulation environments: FetchPush-v1, FetchPickandPlace-v1, Drawer-v1 and Insert-v1. FetchPush-v1 and FetchPickandPlace-v1 are provided in OpenAI Gym \cite{brockman_openai_2016}. These environments are based on a Fetch robotic arm with 7 degrees of freedom (7-DoF) and two parallel grippers. Each state space $s_t$ consists of the kinematic information of the end effector, including coordinates, angular velocity, linear velocity, and gripper open/close status. If there is an object present, the state space also includes the object's coordinates, angular velocity, and linear velocity. Each action space $a_t$ is four-dimensional, where the first three dimensions control the displacement of the end effector and the fourth dimension controls the gripper open/close action. The expected goal is to reach the desired three-dimensional coordinates of the end effector or objects, which is used to determine whether the goal has been achieved. The reward $r_t$ is sparse and binary: a reward of 0 is returned when the goal is achieved within a certain threshold, and a reward of -1 is returned otherwise.

\subsection{Baselines}
To evaluate the performance of MRHER, we compare it with the following baseline algorithms across the four sequential manipulation environments:

\begin{itemize}
\item Deep Deterministic Policy Gradient (DDPG) \cite{lillicrap_continuous_2015} is a model-free off-policy reinforcement learning algorithm for continuous action spaces.
\item Hindsight Experience Replay (HER) \cite{andrychowicz_hindsight_2017} modifies the expected goal of the transition to an achieved goal to explain the action from different goals. In our paper, HER uses the $future$ strategy.
\item Model-based Hindsight Experience Replay (MHER) \cite{yang_mher_2021}, which introduces dynamical models to predict the future trajectory of the current state (with a prediction horizon of 5) by fitting the models to the previous state, action, and next state, and uses it for goal relabeling. 
\item Imaginary Hindsight Experience Replay (IHER) \cite{McCarthy_Wang_Redmond_2021} generates imaginary trajectories by the dynamics models, mixes real trajectories with imaginary trajectories to update the policy and encourages the exploration with an intrinsic reward.
\item Relay Hindsight Experience Replay (RHER) \cite{luo_relay_2022}, which decomposes and recombines a task into subtasks and uses HER with SGES to learn every subtask, achieves state-of-the-art performance in multiple sequential manipulation environments.
\end{itemize}

\begin{figure*}[th]
    \centering
    \includegraphics[width=1\linewidth]{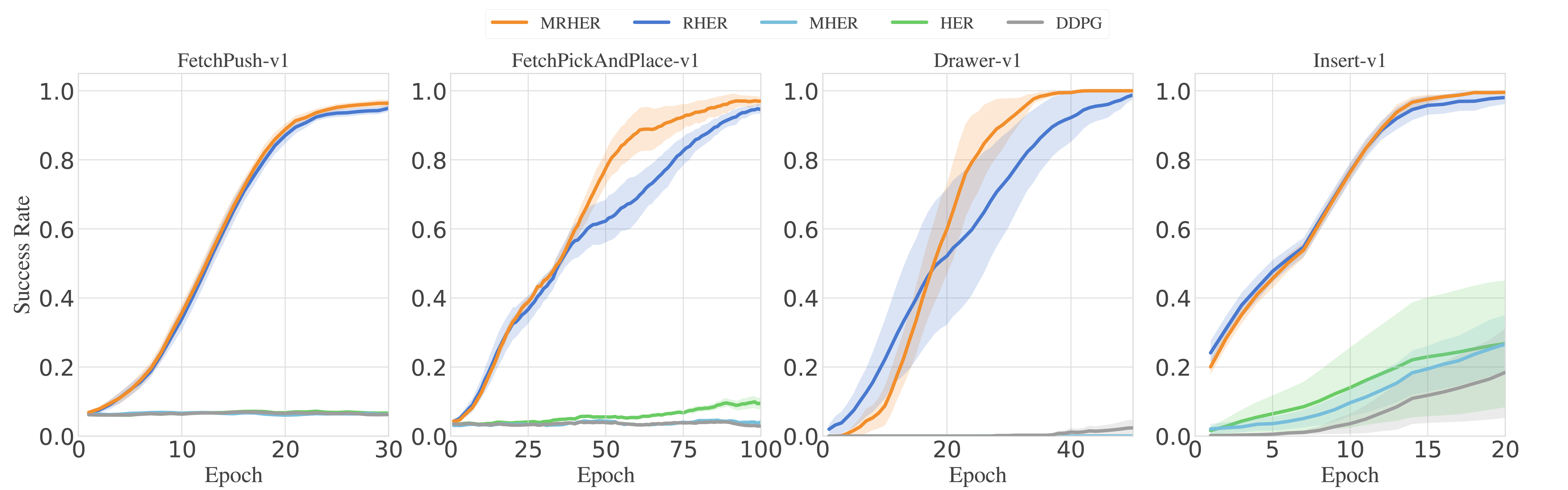}
    \caption{The comparison results between MRHER and baseline algorithms in four sequential manipulation environments.}
    \label{fig:3}
\end{figure*}

\subsection{Training Setting and Evaluation Settings}
According to the difficulty of environments, we train all algorithms within 30 (FetchPush-v1, Insert-v1), 50 (Drawer-v1), or 100 (FetchPickandPlace-v1) episodes on a single CPU core. One core has one rollout worker to generate experience. Each episode consists of 50 trajectories with a interaction step length of 50. After collecting one trajectory, we update all algorithms with 40 batches of size 256 sampled from the replay buffer. During the testing phase, all algorithms are evaluated by executing 50 trials to calculate the average success rate of the current policy after each episode. To obtain reliable results, we conducted the experiments using 5 different random seeds and calculated the average success rate of algorithms.

The dynamics models in MRHER is implemented as a fully connected neural network with 4 hidden layers, each comprising 256 neurons. During the training process, each batch sampled from the replay buffer is used for two updates to the dynamics models. To accelerate exploration, a 0.2 random action noise is added to the interaction with the models $m$ in MHER and MRHER. MRHER uses the same replay ratio as HER, MHER and RHER, which is 0.8. The distance threshold $d$ of MRHER is set to 0.03, matching the threshold used for RHER. The percentage of time a random action is taken is 5\% in MRHER, 20\% in RHER, and 30\% in other baselines.

\subsection{Benchmark Results}

We present the benchmark results of MRHER and baselines in various environments. In MRHER, the step of interaction with the models $m$ is set as 3. Fig.\ref{fig:3} shows the average test success rates of all algorithms in four benchmark environments. The results demonstrate that MRHER outperforms other algorithms by a significant margin. Additionally, Table \ref{table:1} provides additional details regarding the sample efficiency of these algorithms in FetchPush-v1 and FetchPickandPlace-v1. Table \ref{table:1} shows that MRHER demonstrates superior sample efficiency to RHER in FetchPush-v1 and FetchPickandPlace-v1, leading by 13.79\% and 14.29\%, respectively. And the sample efficiency of MRHER is better than that of IHER, which is also a model-based RL algorithm. The results also demonstrate that DDPG and HER learn slowly in all benchmark environments and MHER cannot be effectively trained in most object-manipulation environments. 

\begin{table}[t]
   \caption{Number of interaction steps with real environments to achieve 95\% test success rate with MRHER and baseline algorithms. (1K=1000)}
  \centering
  \begin{tabular}{|c|c|c|c|}
    \hline
    \multicolumn{1}{|l|}{}          & \multicolumn{1}{|l|}{\textbf{FetchPush-v1}} & \multicolumn{1}{|l|}{\textbf{FetchPickandPlace-v1}} \\ {} & 95\% &95\% \\
    \hline
    HER                             & 1170K                                     & 2175K                                             \\ 
    IHER                            & $\sim$80K                                     & $\sim$300K                                             \\ 
    RHER                            & 72.5K                                      & 245K                                              \\ 
    MRHER & \textbf{62.5K}                                      & \textbf{210K}                                              \\ 
    \hline
  \end{tabular}
  \label{table:1}
\end{table}

\begin{figure}[]
    \centering
   \includegraphics[width=1\linewidth]{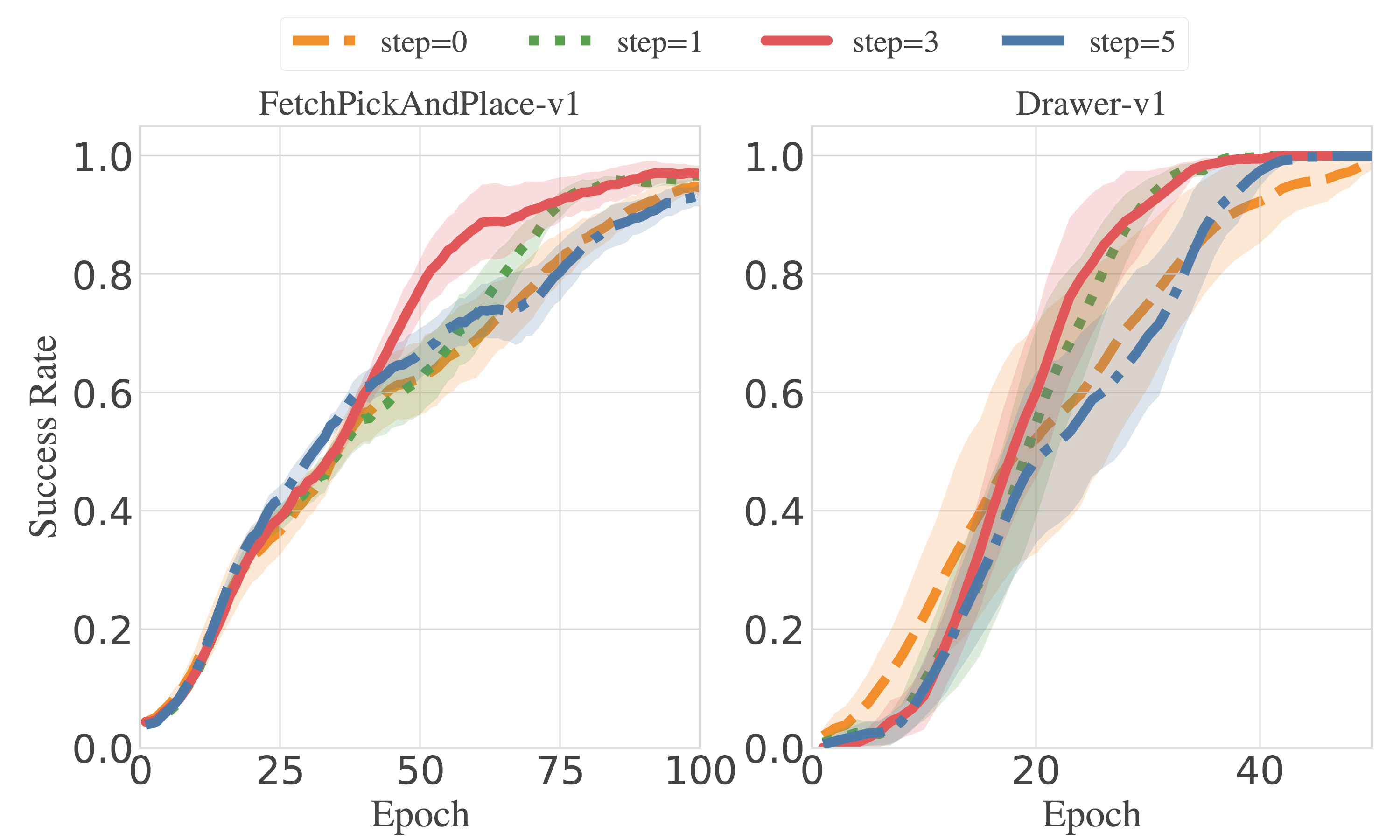}
\caption{Parameter study of step number $n$ in FetchPickandPlace-v1 environment and Drawer-v1 environment.}
\label{fig:4}
\end{figure}

Fig.\ref{fig:4} illustrates the impact of model-based interaction steps $n$ on the performance of MRHER. Different model-based interaction steps $n$ may lead to significant changes in the performance of MRHER, some of which may perform worse. Increasing the model-based interaction steps $n$ allows the agent to interact more with the models and obtain more information. However, too many interaction steps $n$ can result in more accumulated errors and ineffective goals. Fig.\ref{fig:4} reveals that the optimal performance of MRHER is achieved when using 3 interaction steps in FetchPickandPlace-v1 and Drawer-v1.

\begin{figure}[!t]
    \centering
    \includegraphics[width=0.70\linewidth]{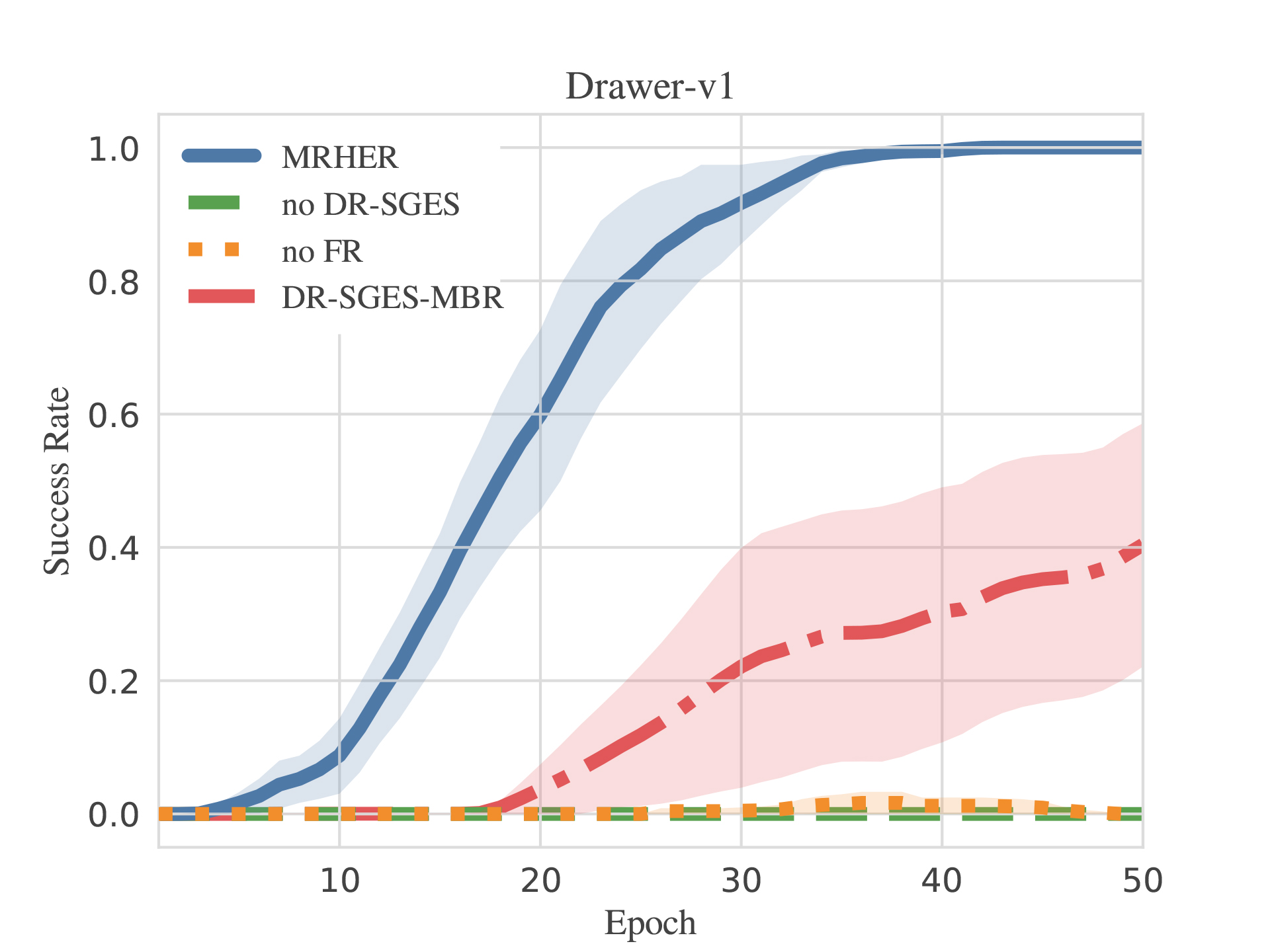}
    \caption{Ablation studies in Drawer-v1 environment.}
    \label{fig:5}
\end{figure}

To analyze the importance of components in the MRHER framework, we conducted experiments on different versions of MRHER. The default step number of interaction with the model $m$ is set to 3. The decomposition and recombination of a sequential task and the SGES are closely related and can be considered as a whole component, named DR-SGES. The experimental settings are as follows:
\begin{itemize}
\item MRHER: DDPG + DR-SGES + FR;
\item no DR-SGES: DDPG + FR;
\item no FR: DDPG + DR-SGES;
\item DR-SGES-MBR: DDPG + DR-SGES + MBR.
\end{itemize}
The empirical results depicted in Fig.\ref{fig:5} demonstrate that both FR and DR-SGES are important components in MRHER framework for alleviating the INNR problem and achieving high success rates. DDPG + FR and DDPG + DR-SGES learn slowly in Drawer-v1. However, MRHER, which combines both DR-SGES and FR components, achieves the highest success rate in the fewest steps, which means that FR can accelerate the learning process of the agent. Additionally, we conducted an ablation study where FR is replaced by MBR in MRHER. Fig.\ref{fig:5} shows that DR-SGES-MBR, which replaces FR with MBR, has lower sample efficiency compared to MRHER with FR. This difference in sample efficiency indicates that FR plays a crucial role in the MRHER framework.

\section{Conclusion}
We propose a novel method called Model-based Relay Hindsight Experience Replay (MRHER) to address the INNR issue of existing model-based goal relabeling methods in sequential manipulation environments. The cores of MRHER are the decomposition and recombination of a sequential task, a self-guided exploration strategy and a novel robust goal relabeling method called Foresight relabeling. Experiments results demonstrate that MRHER achieves state-of-the-art performance in all benchmark environments. For future research, we aim to expand the application of MRHER to other areas of reinforcement learning, exploring its potential in different domains and tasks.

\bibliographystyle{IEEEtran}

\bibliography{main}

\begin{thebibliography}{10}
\providecommand{\url}[1]{#1}
\csname url@samestyle\endcsname
\providecommand{\newblock}{\relax}
\providecommand{\bibinfo}[2]{#2}
\providecommand{\BIBentrySTDinterwordspacing}{\spaceskip=0pt\relax}
\providecommand{\BIBentryALTinterwordstretchfactor}{4}
\providecommand{\BIBentryALTinterwordspacing}{\spaceskip=\fontdimen2\font plus
\BIBentryALTinterwordstretchfactor\fontdimen3\font minus \fontdimen4\font\relax}
\providecommand{\BIBforeignlanguage}[2]{{%
\expandafter\ifx\csname l@#1\endcsname\relax
\typeout{** WARNING: IEEEtran.bst: No hyphenation pattern has been}%
\typeout{** loaded for the language `#1'. Using the pattern for}%
\typeout{** the default language instead.}%
\else
\language=\csname l@#1\endcsname
\fi
#2}}
\providecommand{\BIBdecl}{\relax}
\BIBdecl

\bibitem{durugkar_adversarial_2021}
I.~Durugkar, M.~Tec, S.~Niekum, and P.~Stone, ``Adversarial intrinsic motivation for reinforcement learning,'' in \emph{Advances in Neural Information Processing Systems}, 2021, pp. 8622--8636.

\bibitem{bing_solving_2023}
Z.~Bing, H.~Zhou, R.~Li, X.~Su, F.~O. Morin, K.~Huang, and A.~Knoll, ``Solving robotic manipulation with sparse reward reinforcement learning via graph-based diversity and proximity,'' \emph{{IEEE} Transactions on Industrial Electronics}, vol.~70, no.~3, pp. 2759--2769, 2023.

\bibitem{chen_imitation_2023}
S.~Chen, X.~Ma, and Z.~Xu, ``Imitation learning as state matching via differentiable physics,'' in \emph{Proceedings of the IEEE/CVF Conference on Computer Vision and Pattern Recognition}, 2023, pp. 7846--7855.

\bibitem{luo_relay_2022}
Y.~Luo, Y.~Wang, K.~Dong, Q.~Zhang, E.~Cheng, Z.~Sun, and B.~Song, ``Relay hindsight experience replay: Self-guided continual reinforcement learning for sequential object manipulation tasks with sparse rewards,'' \emph{Neurocomputing}, p. 126620, 2023.

\bibitem{andrychowicz_hindsight_2017}
M.~Andrychowicz, F.~Wolski, A.~Ray, J.~Schneider, R.~Fong, P.~Welinder, B.~McGrew, J.~Tobin, P.~Abbeel, and W.~Zaremba, ``Hindsight experience replay,'' in \emph{Advances in Neural Information Processing Systems}, 2017, p. 5055–5065.

\bibitem{Manela_Biess_2021}
B.~Manela and A.~Biess, ``Bias-reduced hindsight experience replay with virtual goal prioritization,'' \emph{Neurocomputing}, vol. 451, p. 305–315, 2021.

\bibitem{Manela_Biess_2022}
------, ``Curriculum learning with hindsight experience replay for sequential object manipulation tasks,'' \emph{Neural Networks}, vol. 145, p. 260–270, 2022.

\bibitem{zhu_mapgo_2021}
M.~Zhu, M.~Liu, J.~Shen, Z.~Zhang, S.~Chen, W.~Zhang, D.~Ye, Y.~Yu, Q.~Fu, and W.~Yang, ``{MapGo}: {Model}-{Assisted} {Policy} {Optimization} for {Goal}-{Oriented} {Tasks},'' in \emph{International Joint Conference on Artificial Intelligence}, vol.~3, 2021, pp. 3484--3491.

\bibitem{yang_mher_2021}
R.~Yang, M.~Fang, L.~Han, Y.~Du, F.~Luo, and X.~Li, ``{MHER}: {Model}-based {Hindsight} {Experience} {Replay},'' in \emph{Advances in Neural Information Processing Systems}, 2021.

\bibitem{fang_curriculum-guided_2019}
M.~Fang, T.~Zhou, Y.~Du, L.~Han, and Z.~Zhang, ``Curriculum-guided {Hindsight} {Experience} {Replay},'' in \emph{Advances in Neural Information Processing Systems}, vol.~32, 2019.

\bibitem{kuang_goal_2020}
Y.~Kuang, A.~I. Weinberg, G.~Vogiatzis, and D.~R. Faria, ``Goal {Density}-based {Hindsight} {Experience} {Prioritization} for {Multi}-{Goal} {Robot} {Manipulation} {Reinforcement} {Learning},'' in \emph{29th {IEEE} {International} {Conference} on {Robot} and {Human} {Interactive} {Communication}}, 2020, pp. 432--437.

\bibitem{atkeson_comparison_1997}
C.~G. Atkeson and J.~C. Santamaria, ``A comparison of direct and model-based reinforcement learning,'' in \emph{Proceedings of International Conference on Robotics and Automation}, vol.~4, 1997, pp. 3557--3564.

\bibitem{sutton_dyna_1991}
R.~S. Sutton, ``Dyna, an integrated architecture for learning, planning, and reacting,'' \emph{ACM SIGART Bulletin}, vol.~2, no.~4, pp. 160--163, 1991.

\bibitem{nagabandi_neural_2018}
A.~Nagabandi, G.~Kahn, R.~S. Fearing, and S.~Levine, ``Neural {Network} {Dynamics} for {Model}-{Based} {Deep} {Reinforcement} {Learning} with {Model}-{Free} {Fine}-{Tuning},'' in \emph{Proceedings of International Conference on Robotics and Automation}, 2018, pp. 7559--7566.

\bibitem{Feinberg_2018}
V.~Feinberg, A.~Wan, I.~Stoica, M.~I. Jordan, J.~E. Gonzalez, and S.~Levine, ``Model-based value expansion for efficient model-free reinforcement learning,'' in \emph{Proceedings of International Conference on Robotics and Automation}, 2018.

\bibitem{buckman2018sample}
B.~Jacob, H.~Danijar, T.~George, B.~Eugene, and L.~Honglak, ``Sample-efficient reinforcement learning with stochastic ensemble value expansion,'' in \emph{Advances in Neural Information Processing Systems}, vol.~31, 2018.

\bibitem{bai_guided_2019}
C.~Bai, P.~Liu, W.~Zhao, and X.~Tang, ``Guided goal generation for hindsight multi-goal reinforcement learning,'' \emph{Neurocomputing}, vol. 359, pp. 353--367, 2019.

\bibitem{charlesworth_plangan_2020}
H.~Charlesworth and G.~Montana, ``Plangan: Model-based planning with sparse rewards and multiple goals,'' in \emph{Advances in Neural Information Processing Systems}, vol.~33, 2020, pp. 8532--8542.

\bibitem{ghosh_learning_2021}
D.~Ghosh, A.~Gupta, A.~Reddy, J.~Fu, C.~M. Devin, B.~Eysenbach, and S.~Levine, ``Learning to {Reach} {Goals} via {Iterated} {Supervised} {Learning},'' in \emph{International Conference on Learning Representations}, 2021.

\bibitem{schaul_universal_2015}
T.~Schaul, D.~Horgan, K.~Gregor, and D.~Silver, ``Universal {Value} {Function} {Approximators},'' in \emph{Proceedings of {International} {Conference} on {Machine} {Learning}}, 2015, pp. 1312--1320.

\bibitem{mnih_human-level_2015}
V.~Mnih, K.~Kavukcuoglu, D.~Silver, A.~A. Rusu, J.~Veness, M.~G. Bellemare, A.~Graves, M.~Riedmiller, A.~K. Fidjeland, G.~Ostrovski, S.~Petersen, C.~Beattie, A.~Sadik, I.~Antonoglou, H.~King, D.~Kumaran, D.~Wierstra, S.~Legg, and D.~Hassabis, ``Human-level control through deep reinforcement learning,'' \emph{Nature}, vol. 518, no. 7540, pp. 529--533, 2015.

\bibitem{lillicrap_continuous_2015}
T.~P. Lillicrap, J.~J. Hunt, A.~Pritzel, N.~Heess, T.~Erez, Y.~Tassa, D.~Silver, and D.~Wierstra, ``Continuous control with deep reinforcement learning,'' \emph{arXiv preprint arXiv:1509.02971}, 2015.

\bibitem{haarnoja_soft_2018}
T.~Haarnoja, A.~Zhou, P.~Abbeel, and S.~Levine, ``Soft {Actor}-{Critic}: {Off}-{Policy} {Maximum} {Entropy} {Deep} {Reinforcement} {Learning} with a {Stochastic} {Actor},'' in \emph{Proceedings of {International} {Conference} on {Machine} {Learning}}, 2018, pp. 1861--1870.

\bibitem{brockman_openai_2016}
G.~Brockman, V.~Cheung, L.~Pettersson, J.~Schneider, J.~Schulman, J.~Tang, and W.~Zaremba, ``{OpenAI} {Gym},'' \emph{arXiv preprint arXiv:1606.01540}, 2016.

\bibitem{McCarthy_Wang_Redmond_2021}
R.~McCarthy, Q.~Wang, and S.~J. Redmond, ``Imaginary hindsight experience replay: Curious model-based learning for sparse reward tasks,'' \emph{arXiv preprint arXiv:2110.02414}, Oct. 2021.

\end{thebibliography}

\vspace{12pt}
\color{red}
\end{document}